# An Image Clustering Auto-Encoder Based on Predefined Evenly-Distributed Class Centroids and MMD Distance

Qiuyu ZHU, Zhengyong WANG

School of Communication & Information Engineering, Shanghai University

## Abstract

In this paper, we propose a novel, effective and simpler end-to-end image clustering auto-encoder algorithm: ICAE. The algorithm uses PEDCC (Predefined Evenly-Distributed Class Centroids) as the clustering centers, which ensures the inter-class distance of latent features is maximal, and adds data distribution constraint, data augmentation constraint, auto-encoder reconstruction constraint and Sobel smooth constraint to improve the clustering performance. Specifically, we perform one-to-one data augmentation to learn the more effective features. The data and the augmented data are simultaneously input into the autoencoder to obtain latent features and the augmented latent features whose similarity are constrained by an augmentation loss. Then, making use of the maximum mean discrepancy distance (MMD), we combine the latent features and augmented latent features to make their distribution close to the PEDCC distribution (uniform distribution between classes, Dirac distribution within the class) to further learn clustering-oriented features. At the same time, the MSE of the original input image and reconstructed image is used as reconstruction constraint, and the Sobel smooth loss to build generalization constraint to improve the generalization ability. Finally, extensive experiments on three common datasets MNIST, Fashion-MNIST, COIL20 are conducted. The experimental results show that the algorithm has achieved the best clustering results so far. In addition, we can use the predefined PEDCC class centers, and the decoder to clearly generate the samples of each class. The code can be downloaded at https://github.com/zyWang-Power/Clustering!

## 1. Introduction

Clustering is the search for a "natural grouping" in a pile of data. The samples of the same class that we hope to gather are more similar, and the samples of different groups are significantly different. Traditional clustering algorithms, such as [24, 25, 28] classified input data into the same class based on the similarity of extracted features. However, these were general clustering algorithms, not particularly for images, and the effects were not good. In the past, feature extraction and clustering were performed separately and sequentially, such as [7, 35]. But, recent studies by [39, 40, 41] and others have shown that the joint optimization of these two tasks can greatly improve the performance of both. Based on these studies, and along with the development of deep learning, there were many excellent clustering algorithms. [13, 17, 33] used the characteristics of the auto-encoder and some joint training methods to do clustering, the effects were significant. [17] enhanced the clustering effect by maximizing the information of the self-augmented data and the original data. [13] performed better clustering by extracting features from autoencoder and data augmentation whose algorithms were based on the proposed DEC-DA framework and utilized data augmentation in the step of finetuning. Our method is totally different, and data augmentation is used directly in training.

In this paper, we propose a new, effective and simpler end-to-end clustering autoencoder algorithm: ICAE as shown in Figure 1, and verify the algorithm on different datasets, which has achieved the best results so far. Specifically, one-to-one data augmentation is performed before the data is input to the encoder, and the data and the augmented data are simultaneously input to the encoder to obtain latent features and augmented latent features, and we add a constraint loss1 to make them sufficiently similar. Then, we use MMD loss2 for latent features and augmented features to make their distribution approximate to PEDCC distribution (uniform distribution between classes, Dirac distribution within class) to further learn the features used for clustering. At the same time, reconstruction constraint loss3 and Sobel smooth loss4 constraint are added to further improve the generalization ability. In summary, the contributions of our approach are as follows：

- We propose a convolutional neural network, end-to-end unsupervised image clustering auto-encoder algorithm ICAE: using auto-encoder to extract features, PEDCC as cluster centers, using data augmentation constraint, distribution constraint, reconstruction constraint and Sobel smooth constraint to improve performance；
- To the best of our knowledge, these are the first work to use data augmentation directly in clustering training, predefined evenly-distributed class centroids and maximum mean discrepancy in unsupervised deep clustering problem；
- Extensive experiments have demonstrated the effectiveness of our algorithm, which has achieved the best clustering performance on several common datasets；

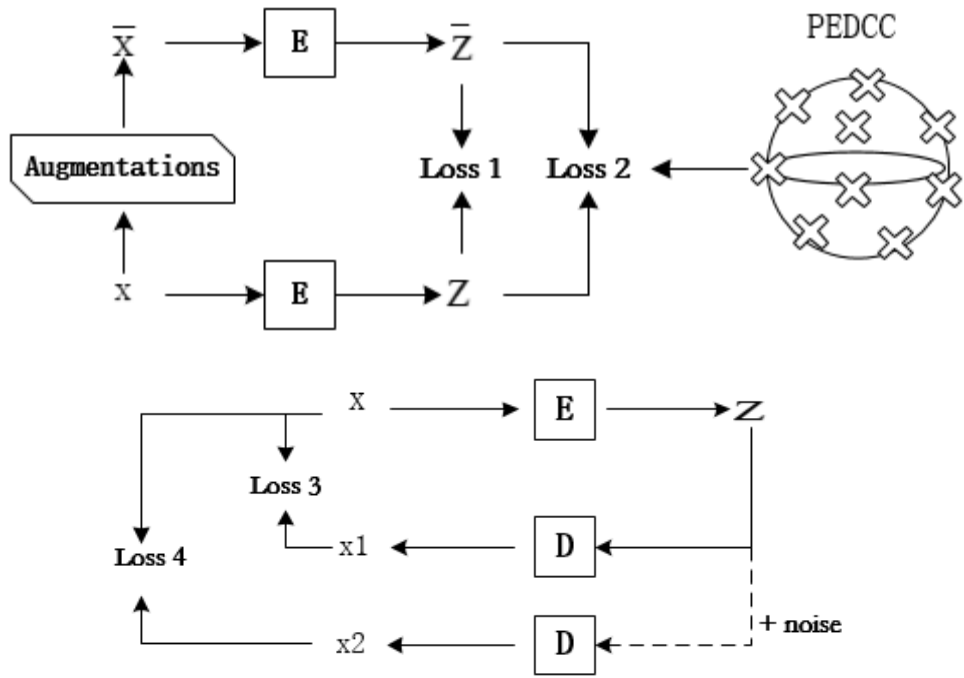

Figure 1 ICAE structure. E and D represent encoder and decoder respectively, and

four different Losses are combined for clustering

This paper is arranged as follows: Related work is summarized in Section 2, and our method is introduced in detail in Section 3. Then, in Section 4, we provide some analysis about computational time and verify the validity, necessity and comparison of the proposed loss designs with the results of MNIST, FASHION, and COIL20 datasets. Finally, Section 5 summarizes the paper and introduces future work.

## 2. Related work

**Clustering:** Existing clustering algorithms can be roughly divided into hierarchical and partitioning methods. [9, 11, 22] were belong to a kind of hierarchical clustering algorithm: starting with many small clusters and then gradually merging into large clusters. The partition clustering method, most notably was K-means [25], minimizes the sum of the squared errors between the data points and their nearest cluster centers. These two similar ideas form the basis of many methods, such as EM [5, 26], SC [28] and so on. In contrast, the clustering centers in our algorithm is pre-defined, satisfies the uniform distribution, the distance between different classes is maximal. Many researchers also used data augmentation to improve clustering effects. For example, [13, 17, 39] used data augmentation to improve generalization performance. But their algorithms utilized self-augmentation or used data augmentation in the step of finetuning. No exception, we also use data augmentation but totally different: we add one-to-one data augmentation directly before input the data to the network, then the data and augmented data are input into the encoder, get the latent features and augmented latent features. We think they should close enough in feature space, so add the constraint to make them similar enough. In Section 3.2, we explore the necessity and effectiveness of data augmentation.

**Auto-encoder for feature extraction:** Autoencoder [8, 20] is an unsupervised learning algorithm that applies backpropagation and sets the target value equal to the input. The autoencoder encoder can learn a valid representation of the input data, called encoding, without any supervision. These codes typically have a much lower dimension than the input data, making the autoencoder available to reduce the number of dimensions. More importantly, autoencoders can act as powerful feature detectors for supervised or unsupervised pre-training of deep neural networks, such as extracting features for human pose recovery and learning 3D faces from 2D images, so on [15, 42, 43]. Finally, they can randomly generate new data that looks very similar to the training data, which is called the generation model. For example, an autoencoder can be trained on a face picture and then a new face can be generated.

The autoencoder works by simply learning to copy the input to the output. It may sound like a trivial task, but limitation of the network in a variety of ways can make it quite difficult. For example, limiting the size of the internal representation, or adding noise to the input and training the network to restore the original input. These constraints prevent the autoencoder from copying the input directly to the output, forcing it to learn an efficient way of representing the data. In short, coding is a by-product of autoencoders trying to learn identity functions under certain constraints.

Many researchers used autoencoders to extract unlabeled data features, such as [13,

17, 33, 39]. [13] made use of denoising autoencoder [36] to extract features, which added noise to the input x and then recovered x from its corrupted version instead of simply copying the input. In this way, the denoising autoencoder can force the encoder and decoder to implicitly capture the structure of the data generation distribution. The difference in this paper is that we do not add noise to the input x, instead we add noise to the latent features after encoding. We construct two losses of the autoencoder: reconstruction loss and Sobel loss. Section 3.4-3.5 explores the necessity and effectiveness of generalized loss design.

**PEDCC (Predefined Evenly-Distributed Class Centroids):** For a conventional deep learning classification network, the strong fitting ability of the neural network can be utilized to make the intra-class distance of the same class small enough. But the distance between different classes is difficult to maximize. The PEDCC [46] approach is to manually set the class centers of latent features so that they are evenly distributed over the surface of the hypersphere to ensure that these cluster centers are farthest away. These class centers are estimated without other unknown variants. We only input the number of categories N and the number of features M. Then making use of the PEDCC algorithm, we can get a group of randomly evenly-distributed class centers on feature hyper-sphere surface. Once determined, these predefined centers will not be changed in the later network training and testing phases.

[46] proposed the Classification Supervised Auto-Encoder (CSAE), which used the predefined uniform distribution class centers to realize the classification function of the autoencoder, and generated different classes of samples according to the class label. But their data were labeled. In contrast, we use PEDCC for clustering. We learn the mapping function through the encoder in the autoencoder and map the different classes of samples to these predefined class centers, so that different classes can be distinguished by the strong fitting ability and effectiveness of deep learning.

## 3. Method

In this section, we will introduce the ICAE algorithm and the losses. Section 3.1 introduces the steps of the ICAE algorithm, Section 3.2-3.5 introduces the design of the loss function, and Section 3.7 introduces the network structure.

### 3.1 ICAE

In this section we will introduce the steps of the ICAE algorithm, see Algorithm 1. We perform one-to-one data augmentation before the data $X$ are input to the encoder to obtain enhanced data $\tilde{X}$, data augmentation operations include rotation, shear, shift, etc. Then the data $X$ and the enhanced data $\tilde{X}$ are input to network simultaneously to obtain the latent features $Z$ and the augmented latent features $\tilde{Z}$. These should be close enough in feature space, so we add constraint loss1 to make them sufficiently similar.

The clustering loss, MMD [21] constraint loss2 are used for latent features and the augmented feature to make their distribution approach to PEDCC distribution to cluster effectively the features. At the same time, reconstruction constraint loss3 and Sobel smooth constraint loss4 are used to improve the generalization ability. The experiments in Sections 3.4 and 3.5 demonstrate that reconstruction constraint and Sobel smooth constraint are necessary and useful.

---

**Algorithm 1**: Image Clustering Auto-Encoder (ICAE) Based on Predefined Evenly-Distributed Class Centroids and MMD Distance

---

**Input**: $X$ = unlabeled images;

**Output**: $k$ classes of clustering centers

**Initialize PEDCC cluster centers;**

**while** Stopping criterion not met **do**

1. $\widehat{X} \leftarrow$ Augumentation($X$);

2. $\widehat{Z} \leftarrow$ Encoder($\widehat{X}$); $Z \leftarrow$ Encoder($X$);

3. Loss 1 = $\text{MSE}(Z, \widehat{Z})$; Loss 2 = $\text{MMD}(Z \odot \widehat{Z}, PEDCC)$ ;

4. $\ddot{X} \leftarrow$ Decoder($Z$ + **noise**); $\dot{X} \leftarrow$ Decoder($Z$);

5. Loss 3 = $\text{MSE}(X, \dot{X})$; Loss 4 = Sobel loss $(X, \ddot{X})$ ;

**end while**

---

### 3.2 Data augmentation loss Function

A lot of works used data augmentation, such as semi-supervised learning and supervised learning. They used data augmentation to enhance the generalization capabilities of the model. The data augmentation methods can be summarized into the following two types:
- Given an input x, the distribution of the output y is determined by x. We can add some noise to the input x or latent features, usually random Gaussian noise to increase the perturbation to improve generalization.
- Set the divergence constraint to minimize the divergence between the dataset and the augmented dataset.

In this work, we use a simpler and more efficient way to increase the perturbation of the dataset to force the autoencoder to extract the more generalized features. Note that, we do not use the augmented data for clustering in the test stage, but use these strategies to get more reliable and realistic data to help encoder extracting the more effective features in training stage. A more reliable and realistic way of data augmentation for a dataset will bring the extracted features closer to the essence of the data.

We believe that compared with traditional disturbances such as adding Gaussian noise, salt and pepper noise, etc. Data augmentation such as rotation, affine transformation, shift, and shear randomly generated one-to-one augmented pictures can be considered a more efficient source of "noise". Specifically, there are several

advantages to use targeted data augmentation as a perturbation function:
- These disturbances are more reliable and realistic disturbances. If an augmented dataset is generated by adding large Gaussian noise to the dataset, the picture may become indistinguishable, or the correct label for the enhanced example may be different from the original example. The purpose of data augmentation is to generate examples that are closer to reality, and use the original examples and augmented examples to share real labels. When we rotate, shear, shift, etc. directly in the original dataset, the label of the augmented image remains the same as the label of the original image. Therefore, the generalization performance of the feature extraction model can be greatly enhanced;
- Targeted data augmentation can be made to different datasets. In the experiment we can directly optimize the data augmentation strategy to improve the performance of each task. For example, the best augmentation strategy on CIFAR-10 primarily involves color-based transformations, such as adjusting brightness, while on MNIST, the best augmentation involves geometric transformations such as shear. These strategies work well;

Suppose a batch-size of samples are $X$, and a one-to-one image set with data augmentation (such as rotation, shear, shift, etc.) are $\tilde{X}$, these operations have a certain range, such as the rotation angle are between (-5, +5). We enter $X$ and $\tilde{X}$ into the encoder separately to get the latent features $Z$ and $\tilde{Z}$. We

$$Loss1(Z,\hat{Z}) = \frac{1}{m}\sum_{i=0}^{m}(Zi - \hat{Z}i)^2 \qquad (1)$$

where **m** is the Batch-Size. Section 4.2 will explore the effectiveness of data augmentation Loss 1.

### 3.3 Clustering loss function

The variational autoencoder [8, 20] imposed a KL divergence [21] constraint on the intermediate layer latent features, bringing it closer to the standard normal distribution. The KL divergence can measure the distance between two distributions, but due to its asymmetry, the application has limitations. Our algorithm uses Maximum Mean Discrepancy (MMD) instead of KL divergence to measure the distance between two distributions. The basic principle of the maximum mean difference is to find a function that assumes that two different distributions have different expectations. If this function is evaluated based on empirical samples in the distribution, this function will indicate whether they are from different distributions. Our loss2 is designed to constrain latent features using the distribution differences between latent features and predefined class centers.

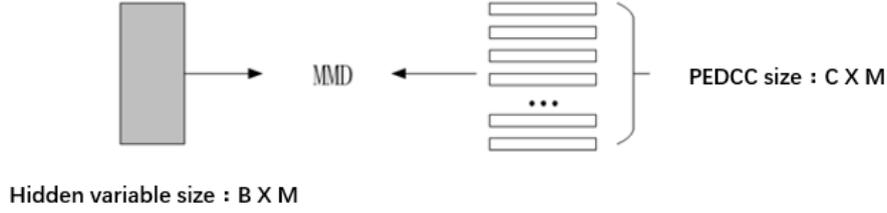

Figure 2 Constraining latent features using maximum mean difference

In Figure 2, the latent features and the augmented latent features are spliced together, and the size of the combined latent features is B×M. Where B is twice of the Batch-Size, which is the sum of the number of images per batch and the number of augmented images. M is the dimension of the latent features. We use the entire batch of image features to compare with a predefined uniform distribution class center, i.e. PEDCC. The size of the predefined uniform distribution class center is C×M, C is the number of predefined evenly distribution class centers, that is, the number of dataset classes, and M is the dimension of each type of predefined uniform distribution centers, which is same as the latent features dimension.

PEDCC is equivalent to some Dirac functions that is uniformly distributed on the normalized latent feature hyper-sphere surface. If the distance between the latent features and the evenly-distributed Dirac functions is close, the intra-class distance after clustering is small. Because the inter-class distance of PEDCC is maximal, the optimal feature distribution can be obtained by minimizing the distance between the distribution of latent features and the PEDCC. Neural network and various loss combinations are used to learn effective features, and MMD can automatically classify the features according to his characteristics. In ideal situation, each class of the cluster is subject to the Dirac distribution. But this is hard to do. So, we can use the MMD (Maximum Mean Difference) to measure the distance between the distribution of latent features of the training samples and the PEDCC. In our approach, MMD loss is the main clustering loss.

We train the network to minimize this distance as the network constraint loss2, namely:

$$loss2 = MMD([Z, \hat{Z}], PEDCC)$$

$$= \frac{1}{M(M-1)} \sum_{i \neq j}^{M} k(l_i, l_j) + \frac{1}{C(C-1)} \sum_{i \neq j}^{C} k(u_i, u_j) - \frac{2}{MC} \sum_{i,j=1}^{M,C} k(l_i, u_j) \quad (2)$$

where $Z$ is the intermediate latent features, $\hat{Z}$ is the latent features of the augmented data, $M$ is its dimension, $l_i = [Z, \hat{Z}]$ is the latent features of the image and its augmented latent features; $U$ represents the PEDCC class centers, $C$ is its number, and $u_i$ is the predefined uniform distribution class center. By iteratively minimizing loss2, it is possible to make the probability distribution of latent features closer to the distribution of the PEDCC. As described above, PEDCC is some points evenly-distributed on the hypersphere. So, the latent features will also approach these points on the hypersphere.

**3.4 Reconstruction loss function**

The autoencoder consists of two parts: an encoder and a decoder. The encoder maps its input x to the representation z in the latent feature space. During training, the decoder attempts to reconstruct x from z, ensuring that useful information are not lost by the encoded phase. In the clustering method, once the training is completed, the decoder portion is no longer used. The encoder is left to map its input into the latent space. By applying this process, the autoencoder can successfully learn useful representations. In order to make the output y of the decoder as close as possible to the input image x, MES of the input image and the output image are directly calculated as the third term of the loss function:

$$loss3 = \frac{1}{m^2}\sum_{i=1}^{m}\sum_{j=1}^{m}(y_{ij} - x_{ij}) \quad (3)$$

where **m** is the image size.

**3.5 Sobel smooth loss**

In addition to data augmentation, we use another approach to improve the generalization of the model. The two complement each other.

CSAE proposed the idea of latent features plus noise, and verified in the paper that the quality of the latent features in the middle layer is greatly improved. We draw on its idea to randomly generate d-dimensional noise $n_0$ from N (0; 1) and add it to the latent features as the input of the decoder. In order to adapt to the increase of the feature space, the network also uses amplifying n0 that satisfies the standard normal distribution.

$$\boldsymbol{n_0^* = \alpha\beta n_0} \quad (4)$$

Where α is the same as the constant coefficient α of PEDCC, and β is a variable between [0, 1]. Therefore, the latent features received by the decoder is a noisy latent feature:

$$\boldsymbol{z_i^* = z_i + n_0^*} \quad (5)$$

However, the addition of noise will deepen the reconstruction of edge blur, and loss3 cannot be solved. Therefore, the Sobel operator is introduced to minimize the difference between the original image and the reconstructed image edge, which is the fourth loss function of the network, loss4.

The Sobel operator consists of two sets of 3*3 matrices, horizontal and vertical, respectively.

$$G_x = \begin{bmatrix} -1 & 0 & 1 \\ -2 & 0 & 2 \\ -1 & 0 & 1 \end{bmatrix}$$

$$G_y = \begin{bmatrix} -1 & -2 & -1 \\ 0 & 0 & 0 \\ 1 & 2 & 1 \end{bmatrix}$$

By flattening it with the image, the difference between the horizontal and vertical brightness differences can be obtained.

First, the gradient strength of each image is obtained after the Sobel operator, and then the average power of them is normalized using L1, that is, the edge positions do not need to correspond exactly, but the average power of the gradient images is the same to maintain edge sharpness.

$$Loss_4(y,x) = \frac{1}{n}|MS(G_x * y_i) + MS(G_y * y_i) - MS(G_x * x_i) - MS(G_y * x_i)| \quad (6)$$

Where $MS(x)$ is the mean square of $x$ (Mean Square).

### 3.6 Loss function

Therefore, combining the above four loss functions, the final loss function of the network is:

$$Loss = a \times Loss_1 + b \times Loss_2 + c \times Loss_3 + d \times Loss_4 \quad (7)$$

where $a$, $b$, $c$, $d$ are the weights of Loss1, Loss2, Loss3, Loss4, respectively, which can be adjusted according to the actual situation.

### 3.7 Network Architecture

In order to make the network have a better fitting ability, the experiment uses a convolutional autoencoder. Specifically, the residual network structure ResNet [14] is used as an encoder, and the deconvolution of the residual network is treated as a decoder. The specific network structure of the decoder and encoder is shown in Table 1. The latent features dimensions of the middle layers, that are, the dimensions of the predefined class centers, the values of the dimensions can be determined experimentally. The following table shows the overall network structure. The output size is exemplified by the MNIST dataset and the dimensions of MNIST latent features are 60.

After the above autoencoder is trained, the trained encoder can be used for feature extraction. For example, the MNIST dataset features are 60-dimensional latent features of the intermediate layer, and the network training makes the intermediate layer latent features approach to the PEDCC, so its characteristics can be well clustered.

Table 1 Network layers of image clustering autoencoder

| layer | Output size | Remarks |
| --- | --- | --- |
| Convolution layer | 32*32 | 32 channels |
| ResNet1 | 16*16 | 64 channels |
| ResNet2 | 8*8 | 128 channels |
| ResNet3 | 4*4 | 256 channels |
| ResNet4 | 2*2 | 512 channels，Encoder output |
| Fully connected layer 1 | 60 | latent features |

| Noise layer | 60 | Gaussian noise |
|---|---|---|
| Fully connected layer 2 | 512*2*2 | |
| ResNet 5（Deconvolution） | 2*2 | 512 channels |
| ResNet 6（Deconvolution） | 4*4 | 256 channels |
| ResNet 7（Deconvolution） | 8*8 | 128 channels |
| ResNet 8（Deconvolution） | 16*16 | 64 channels |
| ResNet 9（Deconvolution） | 28*28 | 32 channels |
| Convolution layer | 28*28 | Rebuild images, one channel |

## 4. Experiments and Discussions

### 4.1 Experimental Setting

**Datasets:** We validate our approach on the following three image datasets. A handwritten digital image dataset MNIST [6], a multi-view object dataset COIL20 [27] and a more concrete human necessity dataset Fashion-MNIST [38], as Table 2. All datasets before input the network is normalized to [-1, 1]. All the experiments are conducted on the same network layers, the parameters are set as Table 3.

Table 2 Datasets

| Dataset | MNIST | COIL20 | Fashion-MNIST |
|---|---|---|---|
| Samples | 70000 | 1440 | 70000 |
| Categories | 10 | 10 | 10 |
| Image Size | 28 * 28 | 128 * 128 | 28 * 28 |

Table 3 Parameters setting of our algorithm

| Hyper-parameter | a | b | c | d |
|---|---|---|---|---|
| Value | 0.01 | 1.00 | 1.00 | 0.05 |

**Experimental Setup**：In the experiment, we keep the network structure unchanged, and determine the value of the super-parameters such as the noise parameter β, the latent features dimension, and the loss function weight. The network is iterated 400 epochs to minimize the loss, and the Adam algorithm is used to optimize the network. The basic learning rate is set to 0.001.

**Evaluation Metrics**：We use the following two indicators to validate our algorithm: Cluster Accuracy (ACC) [34], and Normalized Mutual Information (NMI) [34]. The larger the value, the better the clustering performance.

### 4.2 Analysis on computational time and clustering

We use Pytorch as the deep learning framework and conduct all experiments on an

Inter(R) i7-6700k CPU, 32GB RAM, and a Nvidia GTX 1080 Ti GPU. In the training phase, we construct four loss functions for the encoder and decoder. Convergence is very fast. Taking MNIST as an example, whose training set contains 50,000 pictures, the training time of each epoch is 58 seconds, including the time of enhancing each image in five ways. At 60 epoch we can basically achieve the highest accuracy. In the testing phase, we only need to keep the encoder.

The PEDCC used for training during the training phase are matrix P(N; M), where N is the number of predefined evenly-distribution class centers (that is, the number of dataset classes) and M is the dimension of the latent features, and the latent features are matrix Z(B; M), where B is the Batch-Size, then recognition results are matrix Y(B; N) = $P^T Z$. The row vector values represent the distance of the image to each class. We select the class with the largest distance as the class to which it belongs. Note that, PEDCC and Z are normalized, so the correlation result is the cosine distance. The larger the value, the smaller the angle and the closer it is. We can get the result by correlating the latent features with PEDCC. Only one matrix multiplication is needed, which takes very little time. For 10,000 MNIST test images, only 4 seconds are needed in test stage.

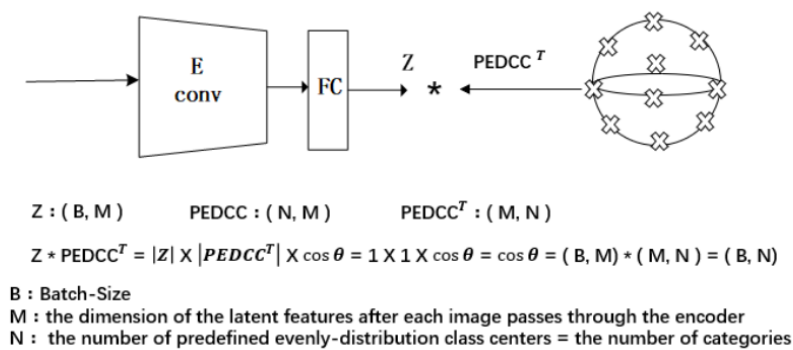

Z : (B, M)     PEDCC : (N, M)     $PEDCC^T$ : (M, N)

$Z * PEDCC^T = |Z| \times |PEDCC^T| \times \cos\theta = 1 \times 1 \times \cos\theta = \cos\theta = (B, M) * (M, N) = (B, N)$

B : Batch-Size
M : the dimension of the latent features after each image passes through the encoder
N : the number of predefined evenly-distribution class centers = the number of categories

Figure 3 The way to do clustering in the test stage

**4.3 Data augmentation effectiveness**

In order to verify the importance of data augmentation, we compare ICAE with data augmentation and ICAE without data augmentation. We set the iteration 400 times, and each specific augmentation are applied randomly for each batch. At the same time, we set the network structure, loss2, loss3 and loss4 unchanged. The difference between data augmentation and without data augmentation are verified on the three datasets. The data augmentation greatly increases the smoothness of the extracted feature model, see Table 4. We can see that ICAE results with data augmentation are better than ICEA without data augmentation, and MNIST has increased by nearly 35%. Even on the more complex Fashion dataset and the 72 images for each type in COIL20 datasets, there is still a significant increase. We found that rotation, shift, and shear work best on MNIST. But the effect on Fashion and COIL20 is not as obvious as MNIST. This inspires us that designing dataset specific data augmentation may further improve clustering performance. We will continue to study general data augmentation.

Let's take the dataset MNIST as an example to elaborate on how to augment data.

Give a set of cluster datasets $X = \{x_i \in R^M\}^n$, where $M$ is the dimension of the latent features and $n$ is the number of sample sets. For each sample $x_i$, we randomly rotate a certain angle $T_{Rotation}$, affine transformation $T_{Affine}$, shear $T_{shear}$, scaling size $T_{scale}$, etc. to get $\widetilde{x_{i1}} = T_{Rotation}(x)$, $\widetilde{x_{i2}} = T_{Affine}(x)$, $\widetilde{x_{i3}} = T_{shear}(x)$, $\widetilde{x_{i4}} = T_{scale}(x)$. We send the original data and the augmented data to the encoder $E(x)$ in order to obtain the latent features: $Z_i$, $Z_{i1}$, $Z_{i2}$, $Z_{i3}$, $Z_{i4}$. We believe that this data augmentation is a very realistic, reliable disturbance, so the features in the feature space should be close enough. So loss1 is specifically represented on the MNIST dataset as:

$$Loss_1(Z, \hat{Z}) = \frac{1}{m} \sum_{i=0}^{m} (Zi - \hat{Z}i)^2$$

$$= \frac{1}{m} \sum_{i=0}^{m} (Zi - Z_{i1})^2 + (Z_{i2} - Z_{i1})^2 + (Z_{i3} - Z_{i2})^2 + (Z_{i4} - Z_{i3})^2 \qquad (8)$$

where m is sample number. For the more complex datasets, Fashion and COIL20, we add translation, pan and flip operation. The proposed data augmentation methods are very simple and effective. Targeted data augmentation for different datasets are more conducive to do clustering.

Table 4 Augmentation experiment results

|  | Data augmentation | Without data augmentation | ACC | NMI |
| --- | --- | --- | --- | --- |
| MNIST | Yes |  | 0.988 | 0.967 |
| MNIST |  | Yes | 0.621 | 0.567 |
| Fashion-MNIST | Yes |  | 0.731 | 0.689 |
| Fashion-MNIST |  | Yes | 0.560 | 0.541 |
| COIL20 | Yes |  | 0.920 | 0.953 |
| COIL20 |  | Yes | 0.660 | 0.741 |

**4.4 Effectiveness of reconstruction constraint**

The encoder turns its input X to the representation Z in the latent feature space and the decoder reconstruct X from Z, ensuring that useful information is not lost by the encoded phase. In order to make the output y of the decoder as close as possible to the input image x, we use MSE loss as a reconstruction constraint. We verified the difference between without reconstruction constraint and with reconstruction constraint on three datasets. As shown in Table 5, we can see that the ICAE with reconstruction constraint is better than the ICAE without reconstruction constraint, and the MNIST has increased by nearly 1%. On the more complicated Fashion-MNIST dataset and COIL20 dataset, there are a bigger improvement, 8% and 2%.

Table 5 reconstruction constraint comparison experiment effect

|  | MSE | Without MSE | ACC | NMI |
|---|---|---|---|---|
| MNIST | Yes |  | 0.988 | 0.967 |
| MNIST |  | Yes | 0.970 | 0.934 |
| Fashion | Yes |  | 0.731 | 0.689 |
| Fashion |  | Yes | 0.656 | 0.644 |
| COIL20 | Yes |  | 0.920 | 0.953 |
| COIL20 |  | Yes | 0.900 | 0.928 |

**4.5 The effectiveness of Sobel smooth constraint**

In order to increase the generalization characteristics of the network. We add the Gaussian noise to the latent features obtained by the raw data input encoder. Gaussian noise satisfies the standard normal distribution. In order to prevent the added noise disturbance from being too large, we set a hyperparameter to limit. In the experiment, the super parameter value was 0.05. We know the introduction of noise by latent features will cause blurred edge of reconstructed images, and *loss*3 cannot be solved entirely. Therefore, the Sobel operator is introduced to minimize the difference between the original image and the reconstructed image edge, which is the fourth loss function of the network, *loss*4. Through experiments, we find that adding a Sobel loss constraint will improve a part of the clustering effect, as Table 6.

Table 6 latent features plus noise and Sobel loss experiment effect

|  | MSE | Without MSE | ACC | NMI |
|---|---|---|---|---|
| MNIST | Yes |  | 0.988 | 0.967 |
| MNIST |  | Yes | 0.979 | 0.946 |
| Fashion | Yes |  | 0.731 | 0.689 |

| | | | | | |
|---|---|---|---|---|---|
| Fashion | | Yes | 0.654 | 0.623 |
| COIL20 | Yes | | 0.920 | 0.953 |
| COIL20 | | Yes | 0.855 | 0.900 |

## 4.6 Comparison with the latest clustering algorithm

We also compare the ICAE algorithm with the most advanced clustering methods available today. In the experiments, we use 60-dimensional dimension of MNIST latent features, and 100 dimensions for Fashion-MNIST because of its complexity. COIL20 has only 72 images for each type, so it uses 160 dimensions. As shown in Table 7, which are the average of 5 trials. In most cases, shallow clustering algorithms such as k-means, spectral clustering, and agglomerative clustering perform worse than deep clustering algorithms. Our ICAE algorithm performs well in the deep clustering algorithm and achieves the best performance on all three data sets. In table 7, all results are reported by running the code they posted or taken from the corresponding paper. The mark " - " means that the result is not available for paper or code.

Table 7 Comparison results with other clustering algorithms

| METHOD | ARCH | NMI MNIST | ACC MNIST | NMI COIL20 | ACC COIL20 | NMI Fashion | ACC Fashion |
|---|---|---|---|---|---|---|---|
| k-means [1] | - | 0.500 | 0.532 | -- | - | 0.512 | 0.474 |
| SC-NCUT [27] | - | 0.731 | 0.656 | - | - | 0.575 | 0.508 |
| SC-LS [28] | - | 0.706 | 0.714 | - | - | 0.497 | 0.496 |
| NMF-LP [29] | - | 0.452 | 0.471 | - | - | 0.425 | 0.434 |
| AC-Zell [30] | - | 0.017 | 0.113 | - | - | 0.100 | 0.010 |
| AC-GDL [31] | - | 0.017 | 0.113 | - | - | 0.010 | 0.112 |
| RCC [32] | - | 0.893* | - | - | - | - | - |
| DCN [8] | MLP | 0.810* | 0.830* | - | - | 0.558 | 0.501 |
| DEC [9] | MLP | 0.834 | 0.863 | - | - | 0.546 | 0.518 |
| IDEC [33] | - | 0.867* | 0.881* | - | - | 0.557 | 0.529 |
| CSC [34] | - | 0.755* | 0.872* | - | - | - | - |
| VADE [35] | VAE | 0.876 | 0.945 | - | - | 0.630 | 0.578 |
| JULE [4] | CNN | 0.913* | 0.964* | - | - | 0.608 | 0.563 |
| DEPICT [36] | CNN | 0.917* | 0.965* | - | - | 0.392 | 0.392 |
| DBC [7] | CNN | 0.917* | 0.964* | - | - | - | - |
| DAC [37] | - | 0.935* | 0.978 | - | - | - | - |
| CCNN [38] | CNN | 0.876 | - | - | - | - | - |
| DEN [39] | MLP | - | - | 0.870 | 0.724 | - | - |
| Neural Clustering [40] | MLP | - | 0.966 | - | - | - | - |
| UMMC [41] | DBN | 0.864 | - | 0.891 | - | - | - |
| TAGNET [42] | - | 0.651 | 0.692 | 0.927 | 0.899 | | |
| IMSAT [12] | MLP | - | 0.983 | - | - | - | - |
| Proposed Method: ICAE | AE | **0.967** | **0.988** | **0.953** | **0.920** | **0.689** | **0.731** |

**4.7 Generated image of ICAE**

We learn the mapping function through the encoder in the autoencoder and map the different classes of samples to these predefined class centers, so that different classes can be distinguished by the strong fitting ability and effectiveness of deep learning. Because the class centers of different classes are already set when they are initialized, we can use PEDCC to generate different classes of images. We can see that all three datasets can accurately generate 10 types of samples as shown in Figure 4, Figure 5, and Figure 6, which also reflects the reliability and accuracy of ICAE clustering. Different classes of images generated by PEDCC :

MNIST（Iteration 40 times）：

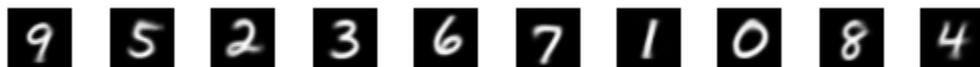

Fashion-MNIST（Iteration 120 times）：

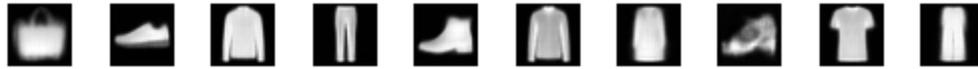
COIL20（Iteration 160 times）

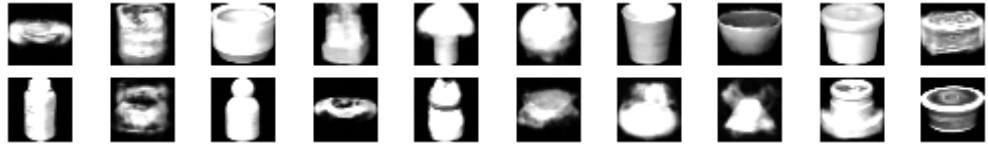

Figure 3 Different classes of images generated by PEDCC

## 5. Conclusions

In this paper, we propose a novel, effective and simpler end-to-end image clustering auto-encoder model: ICAE. The clustering center of the image is set and generated by PEDCC, which ensures the inter-class distance of the latent features is maximal, and the powerful fitting ability of the convolutional neural network are used to map from input images to latent features. Data distribution constraint, data augmentation constraint, autoencoder reconstruction constraint and Sobel smooth constraint are adopted to improve the clustering performance. Extensive experiments show that the algorithm has a good generalization ability, and the best results have been achieved in the three common data sets MNIST, Fashion-MNIST, and COIL20. Our future work will focus on the clustering of some natural images, such as face datasets.